\documentclass[sigplan,nonacm]{acmart}
\usepackage{xspace}

\usepackage[normalem]{ulem}

\usepackage{color}
\usepackage{listings}
\usepackage[]{hyperref}
\usepackage{amsmath,amsfonts}
\usepackage{mathtools}
\usepackage[ruled, vlined, linesnumbered]{algorithm2e}
\usepackage{circledsteps}
\usepackage{comment}
\usepackage{booktabs}
\usepackage{wasysym}
\usepackage{multirow}
\usepackage{graphicx}
\usepackage{makecell}
\usepackage{subcaption}
\usepackage{xcolor}
\usepackage{mathpartir}
\usepackage{adjustbox}
\usepackage{mdframed}
\usepackage{siunitx}
\usepackage{balance}
\usepackage{enumitem}

\usepackage[cachedir=minted-cache]{minted}
\usepackage{tikz}
\usetikzlibrary{tikzmark,fit,backgrounds}

\newcommand{\Name}{Magellan\xspace}
\newcommand{\xae}{AlphaEvolve\xspace}

\begin{document}

\title[\Name: Autonomous Discovery of Novel Compiler Optimization Heuristics with \xae]{\Name: Autonomous Discovery of Novel Compiler\\Optimization Heuristics with \xae}

\author{
Hongzheng Chen$^{1,3}$\enspace
Alexander Novikov$^2$\enspace
Ngân (NV) Vũ$^2$\enspace
Hanna Alam$^1$\enspace
Zhiru Zhang$^3$\enspace\\
Aiden Grossman$^1$\enspace
Mircea Trofin$^1$\enspace
Amir Yazdanbakhsh$^2$\enspace\\
\normalsize $^1$\ Google\enspace $^2$\ Google DeepMind\enspace
$^3$\ Cornell University
}

\renewcommand{\shortauthors}{}

\begin{abstract}
Modern compilers rely on hand-crafted heuristics to guide optimization passes. These human-designed rules often struggle to adapt to the complexity of modern software and hardware and lead to high maintenance burden. To address this challenge, we present \Name, an agentic framework that evolves the compiler pass itself by synthesizing executable C++ decision logic. \Name couples an LLM coding agent with evolutionary search and autotuning in a closed loop of generation, evaluation on user-provided macro-benchmarks, and refinement, producing compact heuristics that integrate directly into existing compilers. Across several production optimization tasks, \Name discovers policies that match or surpass expert baselines. In LLVM function inlining, \Name synthesizes new heuristics that outperform decades of manual engineering for both binary-size reduction and end-to-end performance. In register allocation, it learns a concise priority rule for live-range processing that matches intricate human-designed policies on a large-scale workload. We also report preliminary results on XLA problems, demonstrating portability beyond LLVM with reduced engineering effort.
\end{abstract}

\maketitle

\section{Introduction}
Compilers are among the most mature and systematically engineered software systems, and many of their optimization problems are NP-hard~\cite{dragonbook}, making exact solutions impractical at production scale. As a result, modern compilers rely heavily on hand-crafted heuristics~\cite{fisher1981scheduling,chaitin1982register,lam1988swpipeline,bradlee1991regalloc,chang1989inline}, which encode expert intuition about complex cost-benefit tradeoffs. While these heuristics have enabled decades of progress, they are inherently difficult to design, tune, and maintain. This challenge is only worsening as hardware and software ecosystems become more heterogeneous, eroding the assumptions behind long-standing rules and demanding continual human effort to keep compiler performance competitive.

Recent progress in large language models (LLMs) has renewed interest in automating heuristic design. LLMs have been paired with evolutionary and feedback-driven search to discover algorithms and heuristics for well-defined optimization tasks~\cite{romera2024funsearch,ye2024reevo,liu2024eoh,novikov2025alphaevolve,lange2025shinkaevolve}. Within compilers, most LLM-based efforts fall into two directions: either (1) replacing code generation by directly synthesizing target code~\cite{lange2025cudaengineer,liao2025kernelevolve}, or (2) searching for optimization sequences that drive an existing compiler to better results~\cite{hong2025autocomp,tang2025llmreasoning,cummins2023llmcompiler,cummins2024llmcompiler2}. In contrast, our goal is different. We do not aim to generate per-program pass sequences, nor to bypass the compiler as a code generator. Instead, we target the compiler's \textbf{optimization decision logic} itself by \textbf{evolving the compiler pass code}, producing compact, deployable heuristics that can be integrated upstream and reused across applications.

We argue that this pass-level evolution offers a practical middle ground between manual engineering and neural network (NN) policy integration. Prior NN-based policy discovery methods described in~\cite{trofin2021mlgo,cummins2022compilergym} have shown that learned policies can match or outperform expert heuristics, but they typically require integrating neural models into the compiler infrastructure~\cite{integrateNNs,huang2019autophase,chen2019deeprl}. Reproducing such integration for new passes, new objectives, or emerging accelerators can be expensive and time-consuming. In contrast, our approach treats heuristic design as a program-synthesis problem, directly searching over executable C++ implementations that the compiler can compile and evaluate on realistic macro-benchmarks. As a result, the discovered heuristics retain deployment and maintenance properties that are identical, or very close, to those of human-authored compiler code.

In this paper, we introduce \Name, an autonomous framework for discovering compiler optimization heuristics by combining an LLM-powered coding agent with evolutionary search and autotuning. \Name is designed for real-world, large-scale, production compiler optimization: users can directly use end-to-end application performance metrics as the reward signal, and \Name iteratively proposes, evaluates, and refines a pass implementation that integrates directly into the compiler. We find that even for fundamental, heavily studied optimizations such as function inlining~\cite{theodoros2022optimalinline}, \Name can synthesize new heuristics that \emph{surpass decades of manually engineered effort}, demonstrating that mature compiler passes still contain untapped optimization potential when explored with scalable automated search.

\begin{figure*}[t]
\centering
\includegraphics[width=0.9\linewidth]{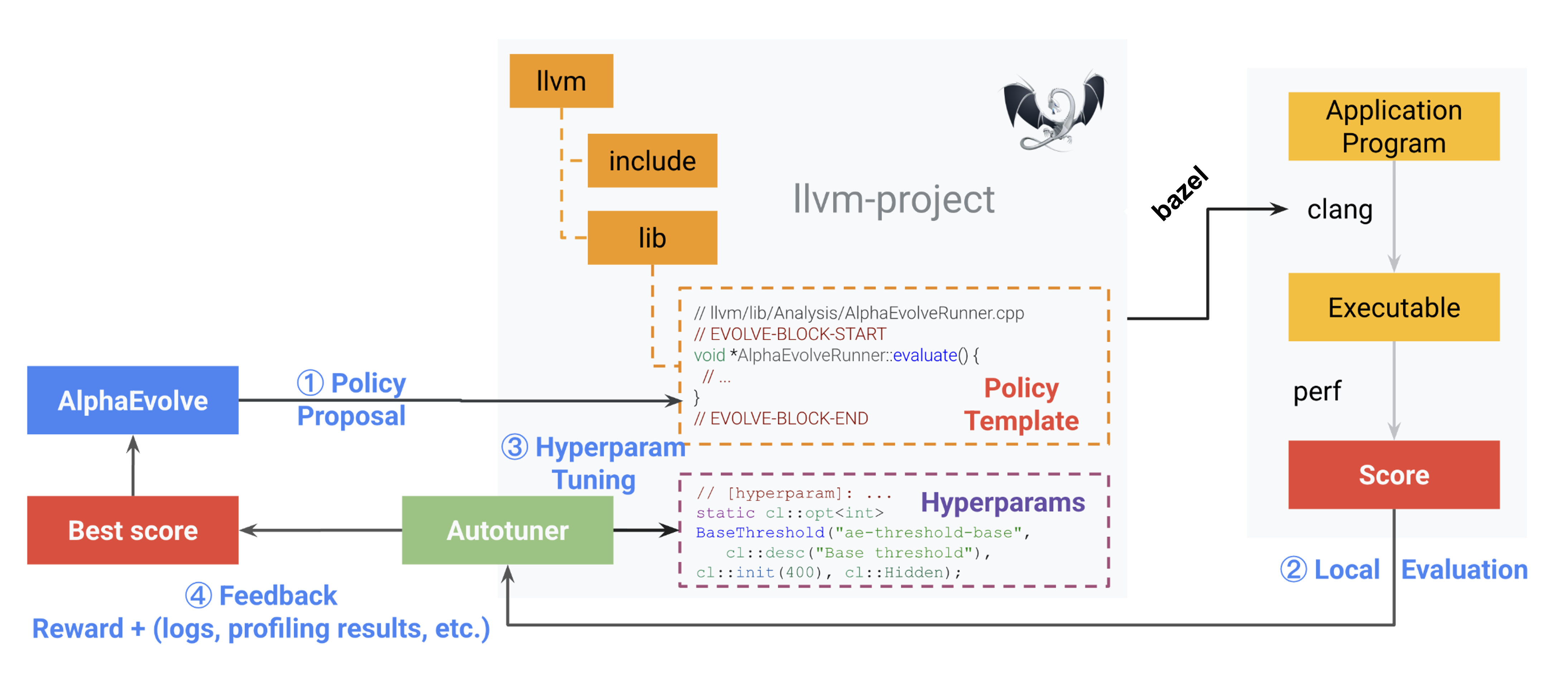}
\caption{Overview of \Name. We use LLVM as the demonstration, but \Name can be readily applied to other compilers.}
\label{fig:overview}
\end{figure*}
We demonstrate the generality and practical impact of \Name through in-depth case studies on heuristic-driven LLVM passes, specifically function inlining and register allocation, and extend our evaluation to additional tasks in XLA. Overall, \Name greatly simplifies automated heuristic discovery in compilers by reducing it to a direct search over human-readable, deployable compiler code, delivering both productivity gains and measurable improvements over expert baselines. Our main contributions are:
\begin{itemize}[leftmargin=*]
\item We introduce \Name, an end-to-end compiler optimization agent powered by \xae~\cite{novikov2025alphaevolve} that evolves compiler passes by synthesizing and evaluating executable C++ heuristics with a simple integration pass into existing compilers.
\item We introduce a hierarchical search strategy in \Name that improves evolution efficiency by combining an LLM to propose heuristic templates with an autotuner to refine hyperparameters during search.
\item We present case studies on realistic compiler optimization tasks in LLVM and XLA, showing code-size reduction and runtime speedups over expert-crafted baselines.
\end{itemize}

\section{\Name Design}

The process illustrated in Fig.~\ref{fig:overview} presents the four-stage workflow of \Name, which integrates \xae with the LLVM compiler to automate the discovery and refinement of optimization heuristics.
LLVM can be substituted with other infrastructures such as XLA to target different compiler passes and backends.
The key idea of \Name is a hierarchical search strategy that \emph{separates high-level policy design from low-level parameter tuning}: the LLM proposes a heuristic template with symbolic hyperparameters, and an external autotuner later fills in those values. The system operates as a closed-loop process comprising \textbf{(1) policy proposal}, \textbf{(2) local evaluation}, \textbf{(3) hyperparameter tuning}, and \textbf{(4) feedback generation}, forming an iterative cycle that gradually improves compiler decision policies.

\subsection{Policy Proposal}
In the first stage, \Name uses \xae equipped with an LLM to modify a source file that contains the policy we want to improve. The policy-relevant regions are marked with \texttt{EVOLVE-BLOCK} comments, a mechanism provided by \xae to delimit editable code. Each candidate defines a decision policy for a compiler optimization pass and conforms to LLVM's standard API interfaces. To structure the search space, we prompt the LLM to emit a policy \emph{template} in which tunable constants are parameterized as explicit compiler flags, allowing the model to focus on high-level decision logic rather than committing to specific numerical thresholds. The synthesized template is then automatically inserted into the LLVM codebase for evaluation.

\subsection{Local Evaluation}
After a candidate heuristic with default hyperparameters is generated, the compiler is recompiled with the modified source file described in the first stage, and the resulting compiler is evaluated on representative macro-benchmarks to obtain an end-to-end reward signal. In our current implementation, we rebuild the compiler and execute the resulting toolchain on the benchmark suite, reporting metrics such as binary size (e.g., via \texttt{llvm-size}) or runtime performance (e.g., via \texttt{perf stat}). This setup grounds evaluation in realistic workloads rather than synthetic objectives, and it also enables us to quantify the sampling complexity required for progress under different configurations. Importantly, \Name is not restricted to full end-to-end execution: when such evaluation is too costly, the same loop can instead rely on proxy signals (e.g., static cost models or faster approximations) as the reward, trading fidelity for scalability.

\subsection{Hyperparameter Tuning}
The resulting performance score is then passed to a box-black autotuner~\cite{golovin2017vizier}, which proposes new hyperparameter configurations based on previous evaluation results. During this stage, the program template remains fixed, while the hyperparameters are iteratively adjusted to seek better performance. After a fixed number of tuning rounds, the best-performing configuration is returned to \xae as feedback for further policy refinement.

\subsection{Feedback Incorporation}
In the final stage, the evaluation results, including the best score, tuning logs, and profiling traces, are returned to \xae. \xae then performs an evolutionary search over policy templates, using these signals together with the empirical reward to select promising candidates and generate new variants for the next iteration. The system then evaluates the new candidate and repeats the loop, iteratively refining the policy template through successive rounds of selection and variation.

\section{Case Studies}
We demonstrate the practicality of \Name by applying it to several key compiler optimization problems. Unless otherwise specified, all experiments use Gemini-2.5-Pro~\cite{gemini25} as the base model. By default, no autotuning is applied. When it is, we use Vizier~\cite{golovin2017vizier} as the autotuner. Our experiments are conducted on Google's internal clang/LLVM toolchain, which closely tracks the LLVM main repository tip-of-tree rather than upstream release snapshots; commits from the 2025 summer timeframe are representative of the version used in our study.

\subsection{Function Inlining for Size}
Function inlining is a classic compiler optimization that replaces function calls with the body of the corresponding callee~\cite{chang1989inline}. This transformation eliminates call overhead and exposes additional optimization opportunities such as constant propagation, dead code elimination, and register allocation across call boundaries. However, inlining may also increase code size and may degrade instruction cache behavior, making the decision of when and what to inline non-trivial. In fact, globally optimal inlining is known to be NP-hard~\cite{theodoros2022optimalinline}, so modern compilers rely on heuristic cost models to balance performance benefits against code growth and resource constraints.

\begin{figure}[t]
\begin{minted}[linenos,
               fontsize=\scriptsize,
               xleftmargin=1.8em,
               escapeinside=||,
               autogobble]{cpp}
|\text{\textcolor{red}{(a) Partial heuristic definition.}}|
// llvm/lib/Analysis/AlphaEvolveRunner.cpp
// EVOLVE-BLOCK-START
void *AERunner::evaluateUntyped() {
    std::unique_ptr<InlineAdvice>
    int64_t CallerUsers =
    *getTensor<int64_t>(FeatureIndex::caller_users);
    int64_t CalleeUsers =
    *getTensor<int64_t>(FeatureIndex::callee_users);
    // ...
}
// EVOLVE-BLOCK-END

|\text{\textcolor{red}{(b) Full heuristic definition.}}|
// llvm/lib/Analysis/AlphaEvolveRunner.cpp
// EVOLVE-BLOCK-START
AEInlineAdvisor::getAdviceImpl(CallBase &CB) {
    bool IsInliningRecommended = false;
    Function *Callee = CB.getCalledFunction();
    Function *Caller = CB.getCaller();
    // ...
}
// EVOLVE-BLOCK-END
\end{minted}
\caption{Evolvable blocks for function inlining.}
\label{fig:inline}
\Description{}
\end{figure}


Specifically, we choose function inlining for binary size reduction as \Name's pilot problem because it provides a clear and deterministic reward signal, and because we can contrast applying \Name as an exact replacement of a neural network policy~\cite{trofin2021mlgo}.
This optimization is important and is used for edge deployments at Google, such as Fuchsia OS, Chrome on Android, as well as cloud infrastructure.
We consider two settings, with examples shown in Fig.~\ref{fig:inline}:
\begin{itemize}[leftmargin=*]
\item[(a)] \textbf{Partial Heuristics}: Feature-based policies defined by extending \texttt{MLModelRunner::evaluateUntyped}, the interface used to embed neural networks~\cite{trofin2021mlgo}. In this case, \Name needs to compose 38 predefined features\footnote{\url{https://github.com/llvm/llvm-project/blob/c878baf1d9a259cf0788ffa1cc5c9d065adcb4c5/llvm/include/llvm/Analysis/InlineModelFeatureMaps.h}} (e.g., call site and caller/callee usage) to form a new policy.
\item[(b)] \textbf{Full Heuristics}: API-level decision logic implemented via \texttt{MLInlineAdvisor::getAdviceImpl}. \Name only receives a \texttt{CallBase} object and must infer which LLVM APIs to invoke to construct the full policy. Note that starting from \texttt{CallBase}, LLVM's IR interfaces allow traversing the surrounding context, including the enclosing basic block and function, any loops containing the call site, and ultimately the entire compilation unit.
\end{itemize}
In both settings, the heuristic returns a binary decision indicating whether the call site should be inlined. The legality of an inlining decision is checked before invoking the heuristic using the existing \texttt{MLInlineAdvisor} infrastructure. This ensures that, for any policy that integrates successfully into the compiler, the compiler's output is always correct.

\begin{figure}[t]
\includegraphics[width=\linewidth]{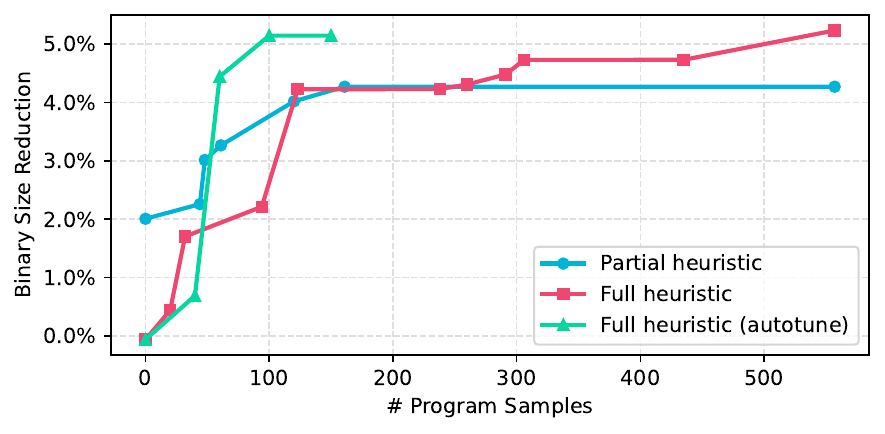}
\caption{Evolutionary curve of binary size reduction on internal search benchmark. Each program sample corresponds to a specific policy template combined with a set of hyperparameters proposed by the autotuner. The initial points differ between the two methods because predefined features influence the inlining decisions prior to search.}
\label{fig:inline_curve}
\end{figure}

We evaluate \Name using an internal search application and start from a naive policy that always returns \texttt{false}. The baseline is the upstream LLVM inlining heuristic\footnote{\url{https://github.com/llvm/llvm-project/blob/main/llvm/lib/Analysis/InlineCost.cpp}}. In the first setting, \Name composes heuristics using the features extracted in the \texttt{MLInlineAdvisor}~\cite{trofin2021mlgo}, intended as inputs for a neural network. In contrast to the previously used neural network-based training techniques~\cite{trofin2021mlgo}, we compute the reward from the final, linked binary, as opposed to compiling random samples of compilation units and using the \texttt{.text} section size in the resulting object file. This feature-based setting exhibits fast early improvements: within a few dozen iterations, the search consistently outperforms the baseline and converges to a stable plateau, achieving a 4.27\% size reduction relative to LLVM's upstream heuristic after 1.5 days of sequential exploration.

In the second setting, \Name generates the entire inlining heuristic directly using LLVM APIs. As shown in Fig.~\ref{fig:inline_curve}, the API-level search starts from a weaker initial point because no predefined features are available to guide decisions, leading to a larger trial-and-error phase in the early stage. However, it continues to improve well beyond the plateau of the feature-based policy, demonstrating the advantage of a more expressive search space. By the end of the 1.5-day search window, the API-based approach yields a 5.23\% improvement over the human-developed heuristic, surpassing both the baseline and the feature-based setting. Notably, the evolutionary curves of the two methods differ not only in final performance but also in learning dynamics: the feature-based search converges faster but saturates earlier, while the API-based search explores more aggressively and achieves higher eventual gains. We provide a detailed analysis of the synthesized policy in Section~\ref{sub:code_dive}.

\subsubsection{Effectiveness of Separation of Concerns}
To evaluate the benefit of separating high-level policy design from low-level parameter tuning, we augment the full-heuristic setting with an external autotuner that searches only over hyperparameters while holding the policy template fixed. In each outer iteration, the autotuner proposes a batch of 10 candidate hyperparameter configurations, and \Name selects the best-performing one for the next iteration. As shown in Fig.~\ref{fig:inline_curve}, this decoupled strategy substantially accelerates convergence and yields higher final gains compared to having an LLM jointly search over both policy logic and hyperparameters. While the untuned full-heuristic setting already outperforms the feature-based setting, adding autotuning further boosts binary size reduction by exploring a richer numeric design space with minimal LLM involvement. Within just 10 outer iterations (100 program samples), \Name achieves more than a 5\% size reduction in roughly 5 hours. This improvement is driven by higher sampling efficiency \emph{by construction}: once a policy template compiles, the autotuner can explore many hyperparameter settings without regenerating the policy, and in particular without repeatedly recompiling the compiler. Consequently, the invalid rate drops from over 65\% when the LLM directly proposes full heuristics to only 13\% with autotuning, yielding a 5$\times$ reduction in wasted samples. Taken together, these results highlight the effectiveness and practicality of separating high-level policy design from low-level parameter tuning.

\subsubsection{Temporal and Domain Generalization}
To understand how well the generated policy generalizes, we evaluate the same policy across both different timestamps and different applications.

\begin{figure}[t]
\centering
\includegraphics[width=\linewidth]{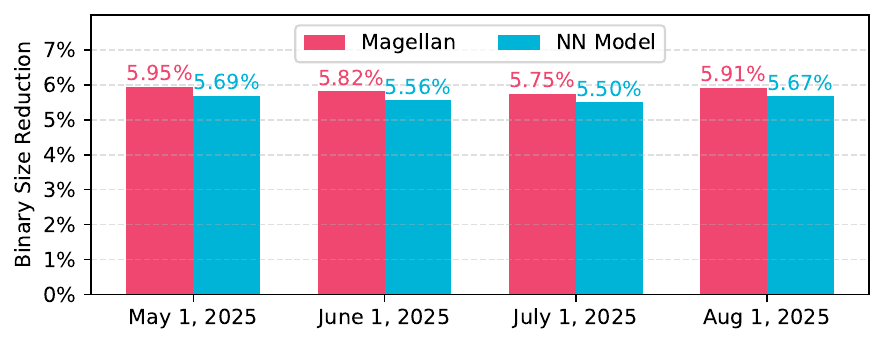}
\caption{Temporal generalization for function inlining.}
\label{fig:temporal}
\end{figure}
As Google uses a monolithic source repository~\cite{monorepo}, a policy is trained at a specific timestamp while the underlying application and codebase evolve over time. We take the best policy produced by \Name in Fig.~\ref{fig:inline_curve} and assess its temporal generalization by measuring performance across four snapshots of the application benchmark collected over consecutive months. We compare the evolved heuristic against the currently-used neural network\footnote{\url{https://github.com/google/ml-compiler-opt/releases/tag/inlining-Oz-v1.2}. It is used by LLVM using the existing, in-tree MLGO facilities, which will not be detailed here.}, which was trained around 2 years ago with a combination of evolutionary strategy~\cite{es2} applied to a large internal code base, followed by further tuning with offline imitation learning~\cite{marinov2024} applied to a relevant internal binary. Despite continuous drift in compiler internals and workload characteristics, \Name consistently delivers superior or comparable results, achieving code size reductions between 5.75\% and 5.95\% and slightly outperforming the neural network. These results suggest that \Name's C++ policies generalize robustly across time without requiring continual retraining or fine-tuning.

\begin{figure}[t]
\centering
\includegraphics[width=\linewidth]{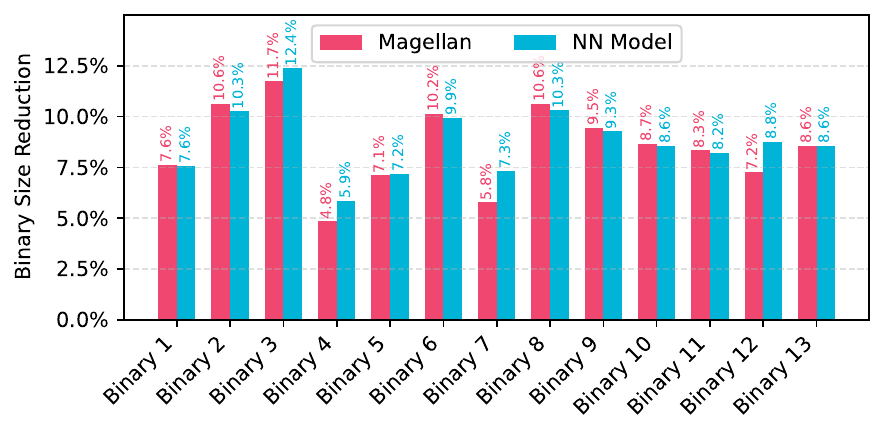}
\caption{Domain generalization for function inlining.}
\label{fig:domain}
\end{figure}
Fig.~\ref{fig:domain} extends the analysis to domain generalization, evaluating \Name across more than ten production binaries compiled with the production optimization flags. Each binary represents a distinct application domain, including search infrastructure and embedded workloads. Here, \Name achieved an average size reduction of 8.79\%, closely matching the neural model's 8.52\%, and outperformed it on a majority of individual binaries. The observed consistency across heterogeneous codebases indicates that the heuristics discovered by \Name capture broadly applicable structural patterns rather than overfitting to specific benchmarks. Moreover, this generalization arises without model retraining or feature re-engineering. Collectively, these results validate that \Name produces heuristics that are both temporally robust and domain generalizable, meeting a criterion for practical deployment in evolving compiler toolchains.

\subsubsection{Code Maintainability}
\label{sub:code_dive}
A detailed analysis of the heuristics synthesized by \Name shows that the generated code is far more compact while achieving superior optimization quality compared to LLVM's manually engineered baseline. The evolved inlining policy contains 143 lines of executable logic as in Appendix~\ref{appendix:inlining}, roughly 15$\times$ shorter than the 2,115-line manual implementation after removing comments and blanks, yet integrates cleanly into LLVM and consistently delivers larger code-size reductions. This conciseness stems from eliminating redundant branching conditions and special-case logic, retaining only the key decisions needed to balance inlining benefits against code growth. We emphasize, however, that compactness should not be conflated with general maintainability: it remains unclear whether a broader, multi-objective policy (e.g., jointly optimizing size and performance with scope comparable to the default heuristic) would remain similarly concise.

In addition, \Name achieves better code size reduction by identifying that strategic inlining can shrink binaries rather than simply avoiding code growth. Its key insight is to aggressively inline single-use functions so that both the call site and the entire function body disappear, and to prioritize readonly functions that enable substantial dead code elimination after inlining. The generated policy also uses a weighted complexity model that penalizes control-flow instructions more heavily, and adds bonuses for constant arguments and exact type matches that indicate simplification opportunities. In contrast to LLVM's conservative strategy of inlining less for size, this policy recognizes that inlining the right functions, particularly single-use and side-effect-free ones, often leads to smaller final binaries by unlocking downstream optimizations.

\subsection{Function Inlining for Performance}
\begin{figure}[t]
\centering
\includegraphics[width=\linewidth]{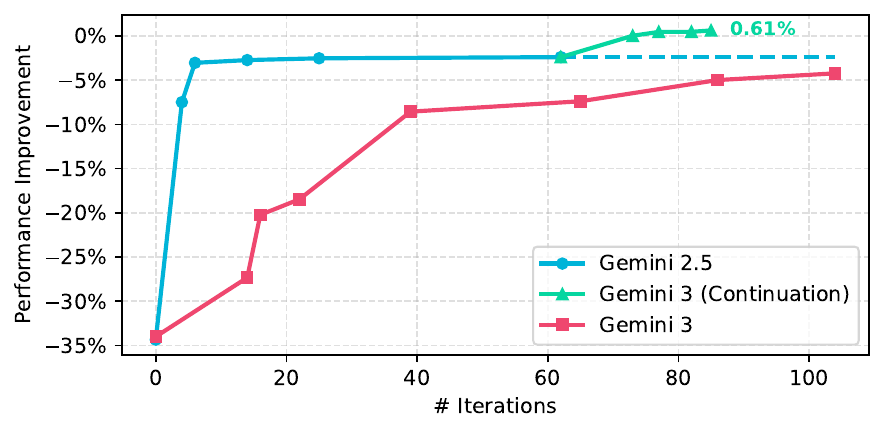}
\caption{Evolutionary curve of performance improvement on the \texttt{clang} benchmark.}
\label{fig:inline4perf}
\end{figure}
We further evaluate \Name on the more challenging task of improving end-to-end performance via function inlining, using LLVM's upstream heuristic as the baseline. As with binary-size reduction, the reward signal is sparse, but performance feedback is both noisy and substantially more expensive to obtain because, beyond recompilation, it requires executing representative \emph{macro}-benchmarks; micro-benchmarks are often poor proxies for large production workloads with flat profiles and diverse hot paths~\cite{kanev2015profiling}. To partially control this cost, we use \texttt{clang} itself as the macro-benchmark, which we judge to be representative while remaining practical to evaluate. The benchmarked binary was built using the same Profile Guided Optimization (PGO) profile, obtained using IR instrumentation, between baseline and experiment, and using distributed ThinLTO~\cite{thinlto} and the -O3 optimization level. We start from a naive policy that returns \texttt{false} for all inlining decisions, and in all experiments the autotuner proposes 40 hyperparameter configurations per iteration. Using Gemini-2.5-Pro as the base model, \Name recovers from the poor starting point and discovers non-trivial improvements, but the search eventually plateaus around a -2.4\% regression after 60 iterations and does not surpass the baseline. We also evaluate Gemini-3-Pro~\cite{gemini3} starting from the same naive \texttt{false} policy, but it similarly fails to push beyond 0\% after 100 iterations (about one week of sequential evaluation), suggesting that starting from scratch, even with a stronger LLM, is insufficient to cross the performance ceiling under this search setting.

To determine whether the plateau reflects a hard performance boundary or a search limitation, we conduct a further experiment: we take the best-performing heuristic produced by Gemini-2.5-Pro and use it as the initialization seed for Gemini-3-Pro, while keeping the same autotuning configuration of 40 hyperparameter proposals per iteration. This continuation strategy constrains the search space around a well-structured heuristic and provides inductive hints that enable Gemini-3-Pro to escape the 0\% plateau. As shown in Fig.~\ref{fig:inline4perf}, the continued Gemini-3-Pro run achieves consistent positive speedups beyond 0\%, ultimately surpassing the hand-tuned baseline by 0.61\%. The evolved policy outperforms LLVM's upstream inliner by using a lightweight cost model with well-tuned heuristics: it skips zero-cost instructions, applies large bonuses for constant arguments, loop nesting, and vector code, and scales thresholds aggressively for hot call sites, inlining the cases that matter most without expensive analysis overhead (Appendix~\ref{appendix:inlining_perf}). To our knowledge, while neural-network policies deployed in LLVM have previously exhibited similar ceiling-breaking behavior~\cite{ashouri2022mlgoperf}, such results have so far been demonstrated only on micro-benchmarks rather than large workloads, and without leveraging (and baselining from) the state of the art in performance optimization, such as PGO and (Thin)LTO~\cite{thinlto}. These findings show that large-model combined with localized hyperparameter tuning can unlock new regions of the performance landscape, highlighting a promising direction for automated performance optimization under sparse and noisy objectives.

\subsection{Preliminary Results on Other Problems}
Beyond function inlining, we also evaluate \Name on additional optimization tasks in LLVM and XLA. The additional task for LLVM is register allocation. In this setting, \Name targets the priority-queue policy used for live-range processing in \texttt{RegAllocGreedy}\footnote{\url{https://github.com/llvm/llvm-project/blob/main/llvm/lib/CodeGen/RegAllocGreedy.cpp}}. On a large-scale production search workload, \Name converges to an unexpectedly simple solution: a constant-value policy, which implies that the order of live-range insertion alone is sufficient to achieve strong performance in this configuration. Despite its simplicity, the evolved policy matches the performance of intricate human-crafted heuristics, improving from -0.55\% to -0.15\% within relatively few iterations and indicating rapid convergence toward a near-optimal behavior.

To demonstrate \Name's applicability beyond LLVM, we further evaluate it within the XLA compiler stack, focusing on graph rewriting and auto-sharding. The graph rewriting task builds on Enzyme-JAX~\cite{arya2025egraph,moses2020enzyme}, which uses equality saturation~\cite{max2021egg} to apply rewrite rules exhaustively and represent many equivalent computation graphs in an e-graph. After saturation, the key challenge is \emph{extraction}: selecting a single optimized graph under a cost model. The e-graph groups equivalent expressions into e-classes, each containing multiple e-nodes whose children reference other e-classes. A valid extraction chooses one e-node per reachable e-class starting from designated roots, ensures that selecting an e-node also selects exactly one representative from each of its child e-classes, and avoids cycles. This problem is NP-hard and is typically addressed with ILP or heuristics~\cite{yin2025eboost,cai2025smoothe}. In our setting, each e-node is assigned a cost, and the objective is to minimize the total cost of the extracted program. \Name synthesizes new extraction heuristics that guide these selection decisions, achieving a 7\% improvement over manually designed policies and demonstrating its ability to navigate the combinatorial tradeoffs inherent to eqsat-based graph optimization.

The second task targets auto-sharding~\cite{chen2024slapo,sami2025partir,zheng2022alpa} in XLA for distributed TPU execution, following the ASPLOS'25 IOPDDL contest setting~\cite{iopddl2025}.
The contest formulates the problem as a constrained combinatorial optimization task over a computational graph: each node must be assigned a sharding strategy from a discrete candidate set, and the objective is to minimize the total cost across both computation and inter-device communication and resharding, while ensuring that time-varying memory usage does not exceed a global budget.
The inputs to this task are graph instances derived from real models (such as diffusion and Gemma), where nodes expose multiple valid strategies and edges capture communication effects between strategy choices. We follow the contest protocol by using the same public benchmark instances for training and the private instances for evaluation (5 public and 20 private in the contest split), and we adopt the same objective and scoring function as the competition. For initialization, we feed the official problem specification\footnote{\url{https://github.com/asplos-contest/2025/blob/main/IOPDDL.md}} to Gemini-2.5-Pro to obtain the initial solution and then apply \Name to evolve the heuristic strategy that selects sharding decisions to reduce the global cost while respecting memory constraints. After one week of evolution, \Name achieves performance comparable to top contest submissions without manually engineered cost models. Although integration into the full XLA pipeline is still in progress by the contest organizers, these results demonstrate that \Name can be applied to large-scale ML compiler optimization problems and is portable beyond LLVM.

\section{Related Work}
Large language models (LLMs) have recently been paired with evolutionary and feedback-driven search to tackle traditional optimization problems with well-defined metrics. FunSearch~\cite{romera2024funsearch} demonstrated that combining an LLM with evolutionary program search can discover improved solutions to traditional combinatorial optimization problems such as cap set and bin packing through iterative program refinement. Subsequent work incorporates LLMs into evolutionary frameworks that use reflection, crossover, and mutation to refine candidate heuristics~\cite{ye2024reevo,liu2024eoh}. AlphaEvolve~\cite{novikov2025alphaevolve} extends this paradigm by evolving entire codebases rather than isolated functions, and demonstrates algorithmic improvements across a range of scientific and engineering applications. ShinkaEvolve~\cite{lange2025shinkaevolve} further advances the approach with techniques that improve sample efficiency and exploration behavior.

The success of these LLM-guided evolutionary frameworks has stimulated interest in system and compiler optimization problems, which often involve combinatorial search spaces. To drive progress in this direction, HeuriGym~\cite{chen2025heurigym} was proposed as one of the first comprehensive suites of system and compiler tasks tailored for LLM evaluation. Broader perspectives on LLM-driven scientific workflows are explored in AI-Driven Research Systems (ADRS)~\cite{cheng2025barbarians}, which highlight the potential for LLMs to automate complex research pipelines across system domains. Task-specific solvers like SATLUTION~\cite{yu2025satlution} apply LLMs to SAT solving.

In LLM-enabled compiler and program generation, prior work largely follows two directions. The first uses the LLM as a \emph{code generator} in place of the compiler, directly emitting target code or accelerator kernels from high-level descriptions~\cite{lange2025cudaengineer,liao2025kernelevolve,hammond2025mtiakernel}. While promising, this line of work is typically limited to relatively small kernels, and it bypasses the mature correctness and compatibility guarantees provided by existing compiler infrastructures. In contrast, we aim to reuse the compiler pipeline to preserve these properties and to target large, production applications. The second direction is more complementary to our goal: instead of generating code, LLMs propose \emph{optimization sequences} or transformation plans that steer an existing compiler toward better outcomes~\cite{hong2025autocomp}. These approaches often pair LLM proposals with structured search such as MCTS~\cite{tang2025llmreasoning} and task-specific fine-tuning~\cite{cummins2023llmcompiler,cummins2024llmcompiler2} to improve decisions in phase ordering and heuristic selection. However, they generally require generating sequences per input application, whereas our approach incurs a one-time cost by directly evolving deployable compiler passes that can be reused across programs.

Complementary to generative approaches, there is a rich literature on machine learning-based heuristic synthesis within compiler infrastructures. Reinforcement learning environments such as CompilerGym~\cite{cummins2022compilergym} enable agents to interact with real compilers to learn optimization sequences from feedback, avoiding per-program generation at inference time. In LLVM specifically, ML-Guided Optimization (MLGO) facilities~\cite{trofin2021mlgo} allow using trained neural policies to replace hand-crafted heuristics for decisions such as inlining or register allocation~\cite{mlgoblog}, demonstrating that learned heuristics can surpass baseline compiler performance without explicit program synthesis.

Our work with \Name aligns with these ML-driven compiler efforts but emphasizes \emph{maintainable and easily deployable heuristic synthesis} by separating high-level policy generation from low-level numeric tuning. \Name synthesizes compact, deployable compiler passes that generalize across applications, combining LLM expressiveness with autotuner efficiency to balance search cost and practical deployment in real-world compiler stacks.

\section{Discussion and Future Work}
The effectiveness of \Name stems from its combination of LLM-driven creativity and systematic evolutionary search over both policy templates and hyperparameters. The LLM proposes a C++ heuristic template that uses LLVM's APIs and exposes its parameters as compiler flags, while the evolutionary search explores sparse reward landscapes in a principled manner. The incorporation of an autotuner that tunes the exposed parameters further improves sampling efficiency. Beyond raw performance gains, a key outcome is a substantial \emph{productivity improvement}: \Name can quickly generate heuristics that match, and often surpass, those developed manually over decades, using the same macro-benchmarks that human developers rely on to obtain an end-to-end reward signal. It produces concise, human-readable heuristic code that can be shipped and maintained just like human-authored compiler code, substantially lowering the barrier to deploying data-driven compiler optimizations in production.

Despite these strengths, several challenges and open questions remain. The current search loop is constrained by the cost of compiling and evaluating each candidate, especially for performance objectives that require evaluating representative macro-benchmarks. It is unclear which properties, if any, make LLM-driven pass evolution competitive (insofar as policy discovery goes) relative to neural-policy approaches, which face similar evaluation cost challenges. Understanding when this alternative is preferable, and how its effectiveness scales with search budget, model choice, and objective noise, is an important direction for future study.

Looking forward, we see several promising directions. First, scaling up evaluations with longer search iterations will help characterize convergence behavior and clarify the limits of the current approach.
Second, \Name can complement neural approaches by automating feature discovery, for example in conjunction with learned embedding methods such as IR2Vec~\cite{venkatakeerthy2020ir2vec}.
Third, \Name should be applied to green-field problems in different compiler infrastructures and domain-specific systems; recent compilers for GPUs and NPUs, such as those targeting kernel mapping, operator scheduling, and task assignment, rely on heuristics that are ripe for automated discovery~\cite{fang2025dato,chen2025tawa,soi2025twill,chen2024allo}. An important question this can answer is whether \Name can discover heuristics in the absence of pre-existing literature and examples.
Finally, we aim for \Name to evolve into a general platform capable of rapidly adapting compiler heuristics to new architectures, optimization tasks, and performance constraints, enabling co-design workflows where humans and AI collaborate fluidly to push the boundaries of automated performance engineering.

\section*{Acknowledgment}
We thank Jacques Pienaar, William Moses, Jeremy Kun, Michael D. Moffitt, Sami Alabed, and Kazu Hirata for providing optimization problems for LLVM and XLA.
We are also grateful to David Li, Tipp Moseley, and Parthasarathy Ranganathan for their feedback on our framework, and to Sagi Perel for his support with Vizier.
We also thank the AlphaEvolve team and the extended team at Google DeepMind who enabled and supported this research direction.

\bibliographystyle{unsrt}
\bibliography{magellan}

\appendix
\onecolumn
\section{\Name-Generated Inlining Policy for Binary Size}
\label{appendix:inlining}
The following code snippet shows the \Name-generated inlining policy for binary size reduction.
\begin{minted}[breaklines,linenos,
               fontsize=\scriptsize,
               xleftmargin=1.8em,
               escapeinside=||,
               autogobble,
               breakanywhere=true,
               fontsize=\scriptsize,
               frame=single]{c++}
// AEInlineAdvisor.cpp
#include "llvm/Analysis/AEInlineAdvisor.h"
#include "llvm/Analysis/InlineAdvisor.h"
#include "llvm/Analysis/OptimizationRemarkEmitter.h"
#include "llvm/IR/PassManager.h"

// EVOLVE-BLOCK-START
// You can include additional LLVM headers here.

using namespace llvm;

// Define constants for inlining thresholds for better readability and maintainability.
// These values are heuristics and can be tuned based on profiling data.
// A very small function, often beneficial to inline to remove call overhead.
constexpr unsigned TINY_FUNCTION_THRESHOLD = 10;
// Base instruction count threshold for regular functions.
constexpr unsigned SMALL_FUNCTION_THRESHOLD = 25;
// Additional instruction count budget for functions with a single call site,
// as inlining them avoids code duplication and enables more optimizations.
constexpr unsigned SINGLE_USE_INLINE_BONUS = 80; // Significantly increases threshold
// Threshold for functions considered to have "many" basic blocks for penalty.
constexpr unsigned MANY_BASIC_BLOCKS_THRESHOLD = 5;
// Penalty applied for functions with many basic blocks.
constexpr unsigned BASIC_BLOCK_PENALTY = 5;

std::unique_ptr<InlineAdvice> AEInlineAdvisor::getAdviceImpl(CallBase &CB) {
  // Implementation of inlining strategy. Do not change the function interface.
  // Default to not inlining.
  constexpr unsigned CONSERVATIVE_INLINE_PENALTY = 20;
  constexpr unsigned HOT_FUNCTION_BONUS = 50;
  bool IsInliningRecommended = false;
  Function *Callee = CB.getCalledFunction();

  // Define weights for different instruction types to calculate a "complexity score"
  // rather than a raw instruction count.
  constexpr unsigned WEIGHT_HIGH_COMPLEXITY_INST = 3;
  constexpr unsigned WEIGHT_MEDIUM_COMPLEXITY_INST = 2;
  constexpr unsigned WEIGHT_LOW_COMPLEXITY_INST = 1;

  // If the callee is null (indirect call) or a declaration (no body to inline),
  // we cannot inline.
  if (!Callee || Callee->isDeclaration()) {
    // If the callee is null (indirect call) or a declaration (no body to inline),
    // we cannot inline. Return advice to not inline.
    IsInliningRecommended = false;
    return std::make_unique<InlineAdvice>(
        this, CB,
        FAM.getResult<OptimizationRemarkEmitterAnalysis>(*CB.getCaller()),
        IsInliningRecommended);
  }

  // Check for explicit inlining attributes first, as they override heuristics.
  if (Callee->hasFnAttribute(Attribute::NoInline)) {
    // Explicitly prevent inlining. IsInliningRecommended remains false.
    // Fall through to the final return.
  } else if (Callee->hasFnAttribute(Attribute::AlwaysInline)) {
    IsInliningRecommended = true;
  } else {
    // Crazy Idea: Calculate a weighted instruction count based on instruction complexity.
    // This gives a more nuanced "size" estimation for inlining decisions.
    unsigned WeightedCalleeInstructionCount = 0;
    for (const BasicBlock &BB : *Callee) {
      for (const Instruction &I : BB) {
        switch (I.getOpcode()) {
        case Instruction::Call:
        case Instruction::Invoke:
        case Instruction::CallBr:
        case Instruction::Ret:
        case Instruction::Br:
        case Instruction::Switch:
        case Instruction::IndirectBr:
          WeightedCalleeInstructionCount += WEIGHT_HIGH_COMPLEXITY_INST;
          break;
        case Instruction::Load:
        case Instruction::Store:
        case Instruction::Alloca:
        case Instruction::GetElementPtr:
        case Instruction::AtomicCmpXchg:
        case Instruction::AtomicRMW:
        case Instruction::Fence:
        case Instruction::FNeg:
        case Instruction::FAdd:
        case Instruction::FSub:
        case Instruction::FMul:
        case Instruction::FDiv:
        case Instruction::FRem:
        case Instruction::FCmp:
        case Instruction::SDiv:
        case Instruction::UDiv:
        case Instruction::SRem:
        case Instruction::URem:
        case Instruction::InsertElement:
        case Instruction::ExtractElement:
        case Instruction::ShuffleVector:
        case Instruction::ExtractValue:
        case Instruction::InsertValue:
          WeightedCalleeInstructionCount += WEIGHT_MEDIUM_COMPLEXITY_INST;
          break;
        default:
          WeightedCalleeInstructionCount += WEIGHT_LOW_COMPLEXITY_INST;
          break;
        }
      }
    }

    // Very tiny functions are almost always beneficial to inline due to
    // eliminating call overhead, regardless of other factors. This acts as
    // an absolute minimum threshold that can override other penalties,
    // ensuring even size-optimized tiny functions are considered if they are small enough.
    // Use the raw instruction count for this absolute threshold as it's simpler and
    // less sensitive to specific instruction mixes for extremely small functions.
    if (Callee->getInstructionCount() < TINY_FUNCTION_THRESHOLD) {
      IsInliningRecommended = true;
    } else {
      // Crazy Idea: Aggressive inlining for pure/readonly functions or functions
      // with a single use that are still relatively small. These often lead to
      // significant simplifications and dead code elimination post-inlining,
      // as they reduce overall code size or enable more optimizations.
      constexpr unsigned AGGRESSIVE_SPECIAL_CASE_THRESHOLD = 150;
      // Slightly more conservative for read-only functions than pure functions,
      // as they might still have complex interactions.
      constexpr unsigned AGGRESSIVE_READONLY_THRESHOLD = 75;

      if ((Callee->hasOneUse() && Callee->getInstructionCount() < AGGRESSIVE_SPECIAL_CASE_THRESHOLD) ||
          (Callee->getMemoryEffects().doesNotAccessMemory() && WeightedCalleeInstructionCount < AGGRESSIVE_SPECIAL_CASE_THRESHOLD) ||
          (Callee->getMemoryEffects().onlyReadsMemory() && WeightedCalleeInstructionCount < AGGRESSIVE_READONLY_THRESHOLD)) {
        IsInliningRecommended = true;
      } else {
        // Heuristic-based inlining for other cases:
        unsigned CurrentThreshold = SMALL_FUNCTION_THRESHOLD;

        // Adjust the base threshold based on general function optimization attributes.
        // Use an if-else if structure to prioritize these flags and prevent
        // conflicting adjustments (e.g., being both OptimizeForSize and Hot).
        if (Callee->hasFnAttribute(Attribute::OptimizeForSize) ||
            Callee->hasFnAttribute(Attribute::MinSize)) {
          // For size-optimized functions, start with a more conservative threshold.
          CurrentThreshold = std::max(0U, CurrentThreshold - CONSERVATIVE_INLINE_PENALTY);
        } else if (Callee->hasFnAttribute(Attribute::Hot)) {
          // For hot functions, allow more aggressive inlining.
          CurrentThreshold += HOT_FUNCTION_BONUS;
        }

        // Functions with a single call site are strong candidates as inlining
        // them avoids code duplication and enables more optimizations.
        // This bonus applies if the single-use function wasn't already covered
        // by the more aggressive special-case inlining above.
        if (Callee->hasOneUse()) {
          CurrentThreshold += SINGLE_USE_INLINE_BONUS;
        }

        // Add a penalty for functions with many basic blocks, indicating complex control flow.
        // This is to avoid inflating the caller's basic block count too much if the callee
        // has many small blocks.
        if (Callee->size() > MANY_BASIC_BLOCKS_THRESHOLD) {
          CurrentThreshold = std::max(0U, CurrentThreshold - BASIC_BLOCK_PENALTY);
        }

        // Crazy Idea Enhancement: Add bonus for arguments that are constants or undef.
        // This indicates potential for significant simplification post-inlining.
        constexpr unsigned CONSTANT_ARG_BONUS_PER_ARG = 10;
        constexpr unsigned UNDEF_ARG_BONUS_PER_ARG = 5;
        // Crazy Idea Enhancement: Add bonus/penalty based on argument type matching.
        constexpr unsigned EXACT_TYPE_MATCH_BONUS_PER_ARG = 7;
        constexpr unsigned POINTER_CASTABLE_TYPE_MATCH_BONUS_PER_ARG = 3;
        constexpr unsigned NON_TRIVIAL_CAST_PENALTY_PER_ARG = 5;

        // Get the DataLayout for type comparisons involving pointers.
        const DataLayout &DL = Callee->getParent()->getDataLayout();

        for (unsigned I = 0, E = CB.arg_size(); I != E; ++I) {
          Value *Arg = CB.getArgOperand(I);
          Type *ArgTy = Arg->getType();

          // Check for constant/undef arguments first.
          if (isa<Constant>(Arg)) {
            CurrentThreshold += CONSTANT_ARG_BONUS_PER_ARG;
          } else if (isa<UndefValue>(Arg)) {
            CurrentThreshold += UNDEF_ARG_BONUS_PER_ARG;
          }

          // Check argument type matching against callee's formal parameter type.
          if (I < Callee->getFunctionType()->getNumParams()) {
            Type *ParamTy = Callee->getFunctionType()->getParamType(I);

            if (ArgTy == ParamTy) {
              CurrentThreshold += EXACT_TYPE_MATCH_BONUS_PER_ARG;
            } else if ((ArgTy->isPointerTy() && ParamTy->isIntegerTy()) ||
                       (ArgTy->isIntegerTy() && ParamTy->isPointerTy())) {
              // Check for no-op pointer/integer casts.
              if (CastInst::isBitOrNoopPointerCastable(ArgTy, ParamTy, DL)) {
                CurrentThreshold += POINTER_CASTABLE_TYPE_MATCH_BONUS_PER_ARG;
              } else {
                CurrentThreshold = std::max(0U, CurrentThreshold - NON_TRIVIAL_CAST_PENALTY_PER_ARG);
              }
            } else if (!ArgTy->isVoidTy() && ArgTy != ParamTy) {
              // For other type mismatches (e.g., different integer widths, float to int),
              // apply a penalty if a cast is required that changes bit patterns.
              // We need to infer the cast opcode to check if it's a no-op bitcast or a conversion.
              Instruction::CastOps Opcode = Instruction::BitCast; // Default to BitCast
              if (ArgTy->isIntegerTy() && ParamTy->isIntegerTy()) {
                Opcode = CastInst::getCastOpcode(Arg, /*SrcIsSigned=*/true, ParamTy, /*DstIsSigned=*/true);
              } else if (ArgTy->isFloatingPointTy() && ParamTy->isFloatingPointTy()) {
                Opcode = CastInst::getCastOpcode(Arg, /*SrcIsSigned=*/false, ParamTy, /*DstIsSigned=*/false);
              }

              if (!CastInst::isNoopCast(Opcode, ArgTy, ParamTy, DL)) {
                CurrentThreshold = std::max(0U, CurrentThreshold - NON_TRIVIAL_CAST_PENALTY_PER_ARG);
              }
            }
          }
        }

        IsInliningRecommended = WeightedCalleeInstructionCount < CurrentThreshold;
      }
    }
  }

  // Single return point for all cases.
  return std::make_unique<InlineAdvice>(
      this, CB,
      FAM.getResult<OptimizationRemarkEmitterAnalysis>(*CB.getCaller()),
      IsInliningRecommended);
}

// EVOLVE-BLOCK-END
\end{minted}

\section{\Name-Generated Inlining Policy for Performance}
\label{appendix:inlining_perf}
The following code snippet shows the \Name-generated inlining policy for performance improvement.
\begin{minted}[breaklines,linenos,
               fontsize=\scriptsize,
               xleftmargin=1.8em,
               escapeinside=||,
               autogobble,
               breakanywhere=true,
               fontsize=\scriptsize,
               frame=single]{c++}
#include "llvm/Analysis/AEInlineAdvisor.h"
#include "llvm/Analysis/LoopInfo.h"
#include "llvm/Analysis/BlockFrequencyInfo.h"
#include "llvm/Analysis/InlineAdvisor.h"
#include "llvm/Analysis/OptimizationRemarkEmitter.h"
#include "llvm/Analysis/TargetLibraryInfo.h"
#include "llvm/IR/PassManager.h"

using namespace llvm;

// This is the best parameter configuration obtained by the autotuner:
// [name: "ae-inline-base-threshold"
// int_value: 200
// , name: "ae-inline-call-penalty"
// int_value: 13
// , name: "ae-inline-const-arg-bonus"
// int_value: 68
// , name: "ae-inline-hotness-mul"
// int_value: 10
// , name: "ae-inline-hotness-shift"
// int_value: 3
// , name: "ae-inline-large-caller-reduction"
// int_value: 21
// , name: "ae-inline-large-caller-threshold"
// int_value: 9207
// , name: "ae-inline-loop-bonus"
// int_value: 12
// , name: "ae-inline-recursion-penalty"
// int_value: 22
// , name: "ae-inline-vector-bonus"
// int_value: 97
// ]

// EVOLVE-BLOCK-START
// You can include additional LLVM headers, constants, and flags here.
#include "llvm/IR/Instructions.h"
#include "llvm/IR/InstIterator.h"

// [hyperparam]: ae-inline-base-threshold, int, 10, 200
static cl::opt<int> BaseThreshold(
    "ae-inline-base-threshold",
    cl::desc("Base instruction count threshold for inlining"),
    cl::init(60), cl::Hidden);

// [hyperparam]: ae-inline-call-penalty, int, 5, 50
static cl::opt<int> CallPenalty(
    "ae-inline-call-penalty",
    cl::desc("Penalty for call instructions in callee"),
    cl::init(15), cl::Hidden);

// [hyperparam]: ae-inline-const-arg-bonus, int, 0, 100
static cl::opt<int> ConstantArgBonus(
    "ae-inline-const-arg-bonus",
    cl::desc("Bonus for each constant argument"),
    cl::init(40), cl::Hidden);

// [hyperparam]: ae-inline-loop-bonus, int, 0, 100
static cl::opt<int> LoopBonus(
    "ae-inline-loop-bonus",
    cl::desc("Bonus for call sites inside loops"),
    cl::init(40), cl::Hidden);

// [hyperparam]: ae-inline-vector-bonus, int, 0, 100
static cl::opt<int> VectorBonus(
    "ae-inline-vector-bonus",
    cl::desc("Bonus if callee contains vector instructions"),
    cl::init(30), cl::Hidden);

// [hyperparam]: ae-inline-hotness-mul, int, 1, 10
static cl::opt<int> HotnessMultiplier(
    "ae-inline-hotness-mul",
    cl::desc("Multiplier for hot call sites"),
    cl::init(3), cl::Hidden);

// [hyperparam]: ae-inline-hotness-shift, int, 0, 15
static cl::opt<int> HotnessShift(
    "ae-inline-hotness-shift",
    cl::desc("Shift amount to determine hotness"),
    cl::init(8), cl::Hidden);

// [hyperparam]: ae-inline-recursion-penalty, int, 0, 100
static cl::opt<int> RecursionPenalty(
    "ae-inline-recursion-penalty",
    cl::desc("Penalty for recursive calls"),
    cl::init(50), cl::Hidden);

// [hyperparam]: ae-inline-large-caller-threshold, int, 1000, 10000
static cl::opt<int> LargeCallerThreshold(
    "ae-inline-large-caller-threshold",
    cl::desc("Caller size threshold for penalties"),
    cl::init(4000), cl::Hidden);

// [hyperparam]: ae-inline-large-caller-reduction, int, 0, 90
static cl::opt<int> LargeCallerReduction(
    "ae-inline-large-caller-reduction",
    cl::desc("Percentage reduction of threshold for large callers"),
    cl::init(20), cl::Hidden);

// Implementation of inlining strategy for performance. Do not change the
// function interface. Default implementation is to not inline.
// Please make sure to use BlockFrequencyInfo and TargetLibraryInfo
// as we are optimizing for performance.
std::unique_ptr<InlineAdvice> AEInlineAdvisor::getAEAdviceImpl(CallBase &CB) {
  Function *Caller = CB.getCaller();
  Function *Callee = CB.getCalledFunction();

  if (!Callee || Callee->isDeclaration()) {
    return std::make_unique<InlineAdvice>(
        this, CB, FAM.getResult<OptimizationRemarkEmitterAnalysis>(*Caller),
        false);
  }

  if (Callee->hasFnAttribute(Attribute::AlwaysInline)) {
    return std::make_unique<InlineAdvice>(
        this, CB, FAM.getResult<OptimizationRemarkEmitterAnalysis>(*Caller),
        true);
  }

  const TargetLibraryInfo &TLI = FAM.getResult<TargetLibraryAnalysis>(*Callee);
  LibFunc LibFn;
  if (TLI.getLibFunc(*Callee, LibFn) && TLI.hasOptimizedCodeGen(LibFn)) {
    return std::make_unique<InlineAdvice>(
        this, CB, FAM.getResult<OptimizationRemarkEmitterAnalysis>(*Caller),
        true);
  }

  int Cost = 0;
  bool HasVector = false;

  for (const auto &I : instructions(Callee)) {
    if (isa<AllocaInst>(I) || isa<PHINode>(I) || isa<BitCastInst>(I) ||
        isa<PtrToIntInst>(I) || isa<IntToPtrInst>(I)) {
      continue;
    }

    if (const auto *CI = dyn_cast<CallBase>(&I)) {
      Cost += CallPenalty;
      if (CI->isIndirectCall()) {
        Cost += CallPenalty;
      }
    } else {
      Cost += 1;
    }

    if (!HasVector && I.getType()->isVectorTy()) {
      HasVector = true;
    }
  }

  int Threshold = BaseThreshold;

  for (const auto &Arg : CB.args()) {
    if (isa<Constant>(Arg)) {
      Threshold += ConstantArgBonus;
    }
  }

  LoopInfo &LI = FAM.getResult<LoopAnalysis>(*Caller);
  int Depth = LI.getLoopDepth(CB.getParent());
  if (Depth > 0) {
    Threshold += LoopBonus * std::min(Depth, 3);
  }

  if (HasVector) {
    Threshold += VectorBonus;
  }

  BlockFrequencyInfo &BFI = FAM.getResult<BlockFrequencyAnalysis>(*Caller);
  uint64_t EntryFreq = BFI.getEntryFreq().getFrequency();
  if (EntryFreq > 0) {
    uint64_t CallSiteFreq = BFI.getBlockFreq(CB.getParent()).getFrequency();
    if (CallSiteFreq > (EntryFreq >> HotnessShift)) {
      Threshold *= HotnessMultiplier;
    }
  }

  if (Caller == Callee) {
    Threshold -= RecursionPenalty;
  }

  int CallerSize = 0;
  for (const auto &BB : *Caller) {
    CallerSize += BB.size();
    if (CallerSize > LargeCallerThreshold) break;
  }

  if (CallerSize > LargeCallerThreshold) {
    Threshold = Threshold * (100 - LargeCallerReduction) / 100;
  }

  bool IsInliningRecommended = Cost < Threshold;

  return std::make_unique<InlineAdvice>(
      this, CB, FAM.getResult<OptimizationRemarkEmitterAnalysis>(*Caller),
      IsInliningRecommended);
}

// EVOLVE-BLOCK-END

std::unique_ptr<InlineAdvice> AEInlineAdvisor::getAdviceImpl(CallBase &CB) {
  // legality check
  auto &Caller = *CB.getCaller();
  auto &Callee = *CB.getCalledFunction();
  auto &ORE = FAM.getResult<OptimizationRemarkEmitterAnalysis>(Caller);

  auto MandatoryKind = InlineAdvisor::getMandatoryKind(CB, FAM, ORE);
  // If this is a "never inline" case, there won't be any changes to internal
  // state we need to track, so we can just return the base InlineAdvice, which
  // will do nothing interesting.
  // Same thing if this is a recursive case.
  if (MandatoryKind == InlineAdvisor::MandatoryInliningKind::Never ||
      &Caller == &Callee)
    return getMandatoryAdvice(CB, false);

  auto IsViable = isInlineViable(Callee);
  if (!IsViable.isSuccess())
    return std::make_unique<InlineAdvice>(this, CB, ORE, false);

  bool Mandatory =
      MandatoryKind == InlineAdvisor::MandatoryInliningKind::Always;

  if (Mandatory)
    return std::make_unique<InlineAdvice>(this, CB, ORE, true);

  return getAEAdviceImpl(CB);
}
\end{minted}

\end{document}